\documentclass[runningheads]{llncs}

\usepackage{eccv}

\usepackage{eccvabbrv}

\usepackage{graphicx}
\usepackage{booktabs}
\usepackage{multicol}
\usepackage{graphicx}
\usepackage{algorithm}
\usepackage[noend]{algpseudocode}
\usepackage{amsmath}
\usepackage{booktabs}

\DeclareMathOperator*{\argmin}{arg\,min}
\usepackage[accsupp]{axessibility}  

\usepackage{hyperref}

\usepackage{orcidlink}

\usepackage[table]{colortbl} 

\newcommand{\approach}{\text{\textit{Track2Act}}}
\begin{document}


\title{Track2Act: Predicting Point Tracks from Internet Videos enables Generalizable Robot Manipulation} 

\titlerunning{Track2Act}

\author{Homanga Bharadhwaj$^{1}$ \and Roozbeh Mottaghi$^{2,*}$ \and Abhinav Gupta$^{1,*}$ \and Shubham Tulsiani$^{1,*}$}

\authorrunning{H. Bharadhwaj et al.}

\institute{$^{1}$Carnegie Mellon University and $^{2}$FAIR at Meta}

\maketitle
\begin{abstract}
We seek to learn a generalizable goal-conditioned policy that enables \emph{diverse} robot manipulation — interacting with unseen objects in novel scenes without test-time adaptation. While typical approaches rely on a large amount of demonstration data for such generalization, we propose an approach that leverages web videos to predict plausible interaction plans and learns a task-agnostic transformation to obtain robot actions in the real world.  Our framework, \approach~ predicts tracks of how points in an image should move in future time-steps based on a goal, and can be trained with diverse videos on the web including those of humans and robots manipulating everyday objects. We use these 2D track predictions to infer a sequence of rigid transforms of the object to be manipulated, and obtain robot end-effector poses that can be executed in an open-loop manner. We then refine this open-loop plan by predicting residual actions through a closed loop policy trained with a few embodiment-specific demonstrations. We show that this approach of combining scalably learned track prediction with a residual policy requiring minimal in-domain robot-specific data enables diverse generalizable robot manipulation, and present a wide array of real-world robot manipulation results across unseen tasks, objects, and scenes. \url{https://homangab.github.io/track2act/} \footnote{$^*$ equal contribution. Correspondence to Homanga B. \texttt{hbharadh@cs.cmu.edu}} 
\end{abstract}

\begin{figure*}[h!]
    \centering
    \includegraphics[width=0.95\textwidth]{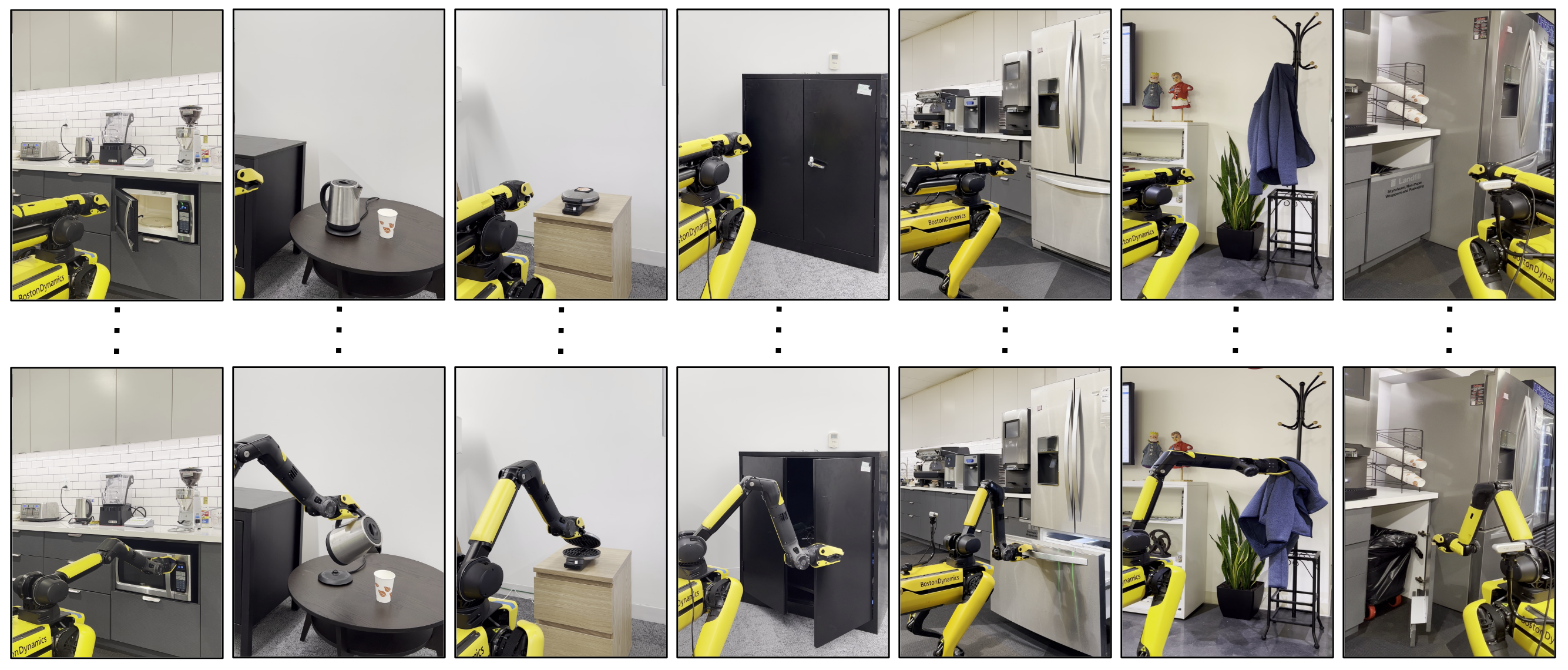}
   \vspace*{-1em}
    \caption{Glimpse of some of the diverse robot manipulation capabilities across physical office and kitchen scenes enabled by our framework. We learn to predict point tracks from web videos for learning interaction plans that can be used for inferring robot actions in unseen scenarios. This enables a \emph{common} goal-conditioned policy to perform everyday tasks like closing microwaves, pulling out drawers, flipping open toasters, pouring from jars etc. Columns show first and last images of rollouts from our policy.} 
    \vspace*{-0.7em}
    \label{fig:teaser}
\end{figure*}
\section{Introduction}

Robots that can be reliably deployed out-of-the-box in new scenarios have the potential for helping humans in everyday tasks. To realize this vision of generalizable robot manipulation, it is crucial to develop \textit{direct} execution capabilities i.e. being able to execute a task out-of-the-box without requiring any test-time training through demonstrations or self-practice before solving a specified task. This is an important desiderata for the system to be repeatedly usable without any downtime, and safe to work alongside humans without performing any exploratory actions.  We pursue the goal of developing such directly executable robot manipulation systems that can perform a broad set of everyday tasks. In addition to being deployable directly, to be widely accessible, we aim to make the robot manipulators generalizable to diverse offices, and kitchens in the real world. 

Developing such directly executable manipulation capabilities has been attempted by prior works, through behavior cloning on robot interaction datasets~\cite{bharadhwaj2023roboagent,rt1,rt2,bcz}. While this approach is in-principle scalable with data, collecting diverse real-world robot interaction data is challenging due to operational constraints. Indeed, recent works that have attempted to scale robot datasets, including cross-robot and cross-domain datasets~\cite{rtx,rh20t,bharadhwaj2023roboagent} still suffer from task diversity issues and are mostly restricted to lab-like structured scenarios. Instead of learning a single-policy that can be directly deployed, some recent works aimed at in-the-wild deployment have adopted the method of test-time training~\cite{bahl2022human,mahi2023bringing}. They require either a video of a human performing the task~\cite{bahl2022human} followed by online exploration, or a demo through a robot end-effector held by a human~\cite{mahi2023bringing}. These approaches are not very convenient for diverse deployments because they require a human to solve the task first, and several hours after that for the robot to learn how to solve that exact task in the exact scene. Thus, such approaches are not directly deployable for new tasks in new scenes.

Our insight to develop an in-the-wild manipulation strategy that is also directly deployable is to factorize a manipulation policy into an \textit{interaction-plan} that can leverage diverse large-scale video sources on the web of humans and robots manipulating everyday objects and a residual policy that requires a small amount of embodiment-specific robot interaction data. Such a factorized structure is inspired by prior works (e.g.~\cite{bharadhwaj2023towards}), however, unlike hand-object masks in~\cite{bharadhwaj2023towards}, we instantiate this \textit{interaction-plan} in an embodiment agnostic-manner by predicting how points in an image of the initial scene move in future frames. This choice of an \textit{interaction-plan} is more expressive compared to hand-object masks adopted by Bharadhwaj et al.~\cite{bharadhwaj2023towards}, as it directly captures point correspondences across time, while at the same time being easier to compute than full RGB frames~\cite{unipy}. Given an initial image of the scene, a goal image defining the task to be performed, and a random set of points in the initial image, we define the interaction plan to be a 2D trajectory of the locations of the points in future frames, such that the goal is achieved. Importantly, we can train this model purely from the abundantly available human and robot videos on the web without any data  specific to the deployment robot embodiment, by using off-the-shelf point-tracking approaches~\cite{cotracker} for generating the ground-truth point trajectories. For deployment in a robot's environment, we can convert the 2D interaction-plan to a sequence of 3D end-effector poses, by having a depth image of the initial scene as an additional input and solving an optimization problem to obtain rigid transforms of the object being manipulated. Finally, with a small amount of embodiment-specific robot interaction data for different tasks ($\sim 400$ trajectories overall), we can learn a goal-conditioned residual policy that corrects for errors in the predicted plan at each time-step and allows for closed-loop deployment.

In summary, we develop \approach~ with the following contributions:
\begin{itemize}
    \item We develop a framework for predicting embodiment-agnostic \textit{interaction-plans} in the form of point tracks from diverse web videos. 
    \item We show how the interaction-plan prediction model can be used for obtaining 3D rigid transforms in a robot's environment for direct manipulation without using any robot data or online exploration.
    \item Given a few ($\sim 400$) embodiment-specific task demonstrations, we show how to learn a goal-conditioned residual policy that can correct for errors in the predicted plan at each time-step. The interaction-plan prediction model combined with the residual policy correction can then be used for closed-loop deployment for new tasks in new scenes.
\end{itemize}

Our real-world robot manipulation results with a Spot robot (highlighted in Fig.~\ref{fig:teaser}) show broad generalization across diverse tasks involving unseen objects in unseen scenes, and demonstrate the potential for leveraging easily available passive videos on the web for learning embodiment-agnostic interaction plans. This is significant as it enables robot manipulation with a \textit{common} goal-conditioned policy, that generalizes to unseen tasks without requiring collection of large scale in-domain manipulation datasets.

\begin{figure*}[t]
    \centering
    \includegraphics[width=\textwidth]{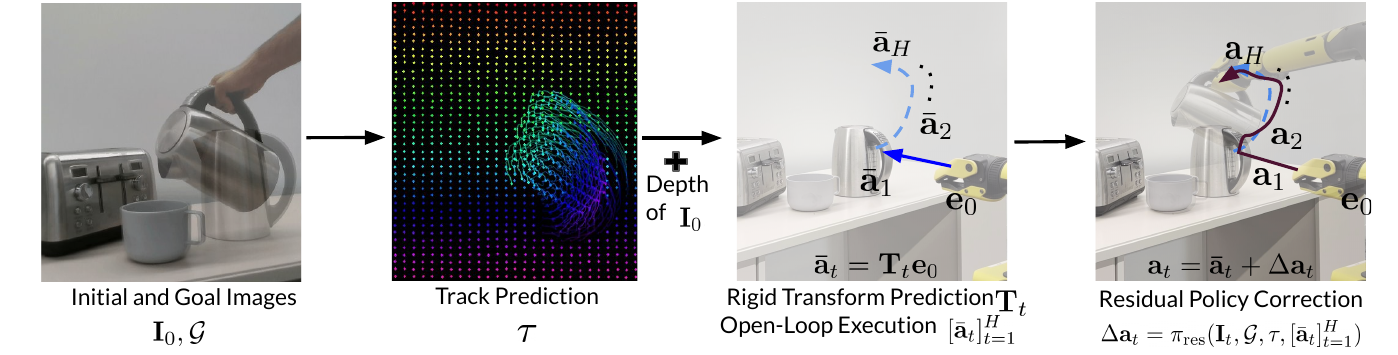}
          \vspace*{-1em}
    \caption{Illustration of the pipeline for learning track prediction from web video datasets, inferring rigid transforms of objects based on the predicted tracks in a robot's environment, and fine-tuning with a residual policy learned with limited robot data. This approach allows us to learn a single goal-conditioned policy for diverse (unseen) tasks.}
    \label{fig:overview}
        \vspace*{-2em}
\end{figure*}

\section{Related Works}
\vspace{-0.2em}

\noindent\textbf{Understanding interactions from videos.} Several computer vision methods have investigated deciphering interactions between hands and objects across various daily activities through curation of large-scale video datasets~\cite{smthsmth,youcook,cmudata,ego4d,Shan20,epic,egtea}, hand pose estimation~\cite{zimmermann2017learning,iqbal2018hand,spurr2018cvpr,ge20193d,baek2019pushing,boukhayma20193d,hasson2019learning,dkulon2020cvpr,liu2021semi,rong2020frankmocap}, 
object pose estimation~\cite{Kehl2017SSD6DMR,Rad2017BB8AS,Xiang2018PoseCNNAC,Hu2019SegmentationDriven6O,He2020PVN3DAD}, interaction hotspot/grasp prediction~\cite{nagarajan2019grounded,liu2022joint,handsasprobes,mo2021where2act,brahmbhatt2019contactgrasp}. Some approaches have investigated \textit{generating} videos given a description of the task, and often conditioned on a scene~\cite{videowilson,unipy,videogen1,unisim}. Other works have attempted understanding generic videos by identifying visual correspondences between frames~\cite{flow1,flow2}. Recent approaches have made significant advances on this problem by developing general-purpose video tracking systems that can track points specified in a frame, across other frames in the video~\cite{tapir,cotracker}. Our track prediction model is inspired by these developments, and is based on leveraging the video tracking approaches to generate ground-truth tracks from web videos, and training a model to \textit{predict} future points tracks given an initial image and a goal.

\noindent\textbf{Learning Visual Representations for Manipulation.}Visual imitation is a promising technique for generalizable robot manipulation~\cite{visual_imitation1,visual_imitation2,visual_imitation3}. Recent works that have scaled this approach for learning large-scale models for manipulation require extremely high number of expert robot trajectories, often demanding years for collection~\cite{rt1,rt2,bharadhwaj2023roboagent}, and still suffer from limited generalization to unseen scenarios for novel objects. Going beyond image observations, prior works have also investigated structured representations like point-clouds~\cite{seita2023toolflownet,taxpose,se3net} and keypoints~\cite{keypoint} for manipulation, but are restricted to tasks in structured table-top scenarios. Some of these that predict action in the form of flow-based representations~\cite{seita2023toolflownet,flow1} require 3D datasets of robot interactions (often from simulation) which constrain them from generalizable real-world deployments. More recently, Vecerik et al.~\cite{vecerik2023robotap}  use point tracking for visual servoing, and the setup requires structured multi-stage definitions of the task and is limited to only minor test-time variations compared to training data. Concurrent work~\cite{wen2023any} that improves upon~\cite{vecerik2023robotap} by predicting future tracks of points in the current image can learn a policy by combining in-domain human videos with in-domain robot videos. However, the framework is not directly amenable for leveraging web videos because the policy relies on per-step image observations for track prediction. Compared to this, and developed independently, we learn to predict trajectories of arbitrary points from web videos given just an initial image and a goal, we show how we can use these predicted tracks to infer rigid transforms of objects for open-loop execution, and further improve the open-loop plan by predicting residuals over the actions, for closed-loop deployment. This enables much diverse robot manipulation behaviors with a single model, that generalizes to unseen novel objects and scenes in-the-wild.

\noindent\textbf{Leveraging Non-Robot Datasets for Manipulation.}  One common way of using data beyond robot interactions for efficient learning is to pre-train the visual representations which serve as backbones for the policy models~\cite{r3m,vip,majumdar2023we,pvr,pvr2} with passive human videos~\cite{ego4d,kay2017kinetics} and image data~\cite{imagenet}. However, these methods still crucially rely on a lot of in-domain robot data or deployment-time training, and are  restricted to learning task-specific policies. Some works that do not require deployment-time training, go beyond visual representations and use curated data of human videos to leverage human hand motion information~\cite{qin2021dexmv,shawvideodex} for learning task-specific policies (instead of a single model across generic tasks). Others that train a single policy across tasks require large in-domain perfectly aligned human-robot data~\cite{wang2023mimicplay,smith2019avid,xiong2021learning} and are not capable of leveraging web data.  Towards learning structure more directly related to manipulation from web videos, some works try to predict visual affordances in the form of where to interact in an image, and local information of how to interact~\cite{mo2021where2act,handsasprobes,bahl2023affordances,liu2022joint}. While these could serve as good initializations for a robotic policy, they are not sufficient on their own for accomplishing tasks, and so are typically used in conjunction with online learning, requiring several hours of deployment-time training and robot data~\cite{bahl2022human,bahl2023affordances}. Others learn to predict masks of hand and objects in the scene~\cite{bharadhwaj2023towards} for conditional behavior cloning and are unable to leverage information of accurate object state changes that is usually ambiguous with a mask. Our work differs from these in terms of predicting an approximate motion of how objects in the scene move in the future through point tracks for the entire trajectory and is directly executable in terms of not requiring any deployment-time training.

\section{Approach}

We aim to develop a generalizable robot manipulation system that can scalably leverage diverse video data for generalizable real-world manipulation. Our key insight (Fig.~\ref{fig:overview}) is to have a factorized policy for 1) learning embodiment-agnostic \textit{interactions plans} of how points in an image of a scene should move in subsequent time-steps to realize a specified goal, followed by 2) inferring robot actions based on the interaction plan through a residual policy. 
We show how this approach allows us to generalize to diverse scenarios involving unseen tasks and objects, since the prediction model by virtue of being trained on web data generalizes well to new scenes, and the residual policy has a much simpler task of correcting the robot actions derived from the interaction plan. 

\begin{figure}[t]
    \centering
    \includegraphics[width=0.95\columnwidth]{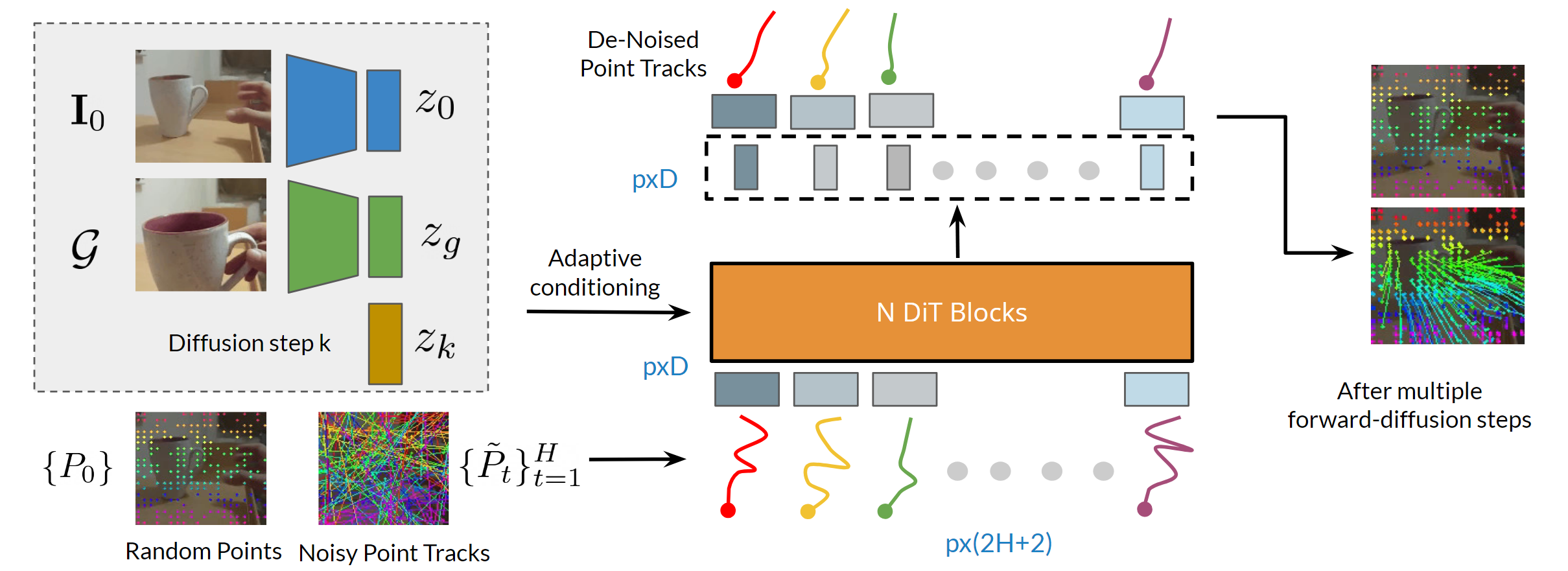}
    \caption{Architecture of the Diffusion Transformer $\mathcal{V}_\theta$ for denoising track predictions given initial image $\mathbf{I_0}$, goal $\mathcal{G}$, and an initial set of $p$ points $P_0$.}
    \label{fig:arch}
\end{figure}

\subsection{Overview and Setup}
Given a scene specified by an RGB image $\mathbf{I_0}$ and a goal image $\mathcal{G}$ denoting what task should be performed, we want to have a robot manipulator execute actions $\mathbf{a}_{1:H}$ in the scene to achieve the desired goal. To achieve this in unseen scenarios, we leverage web video data by learning a model $\tau = \mathcal{V}_\theta(\mathbf{I_0},\mathcal{G},P_0)$ to predict future locations (tracks) of $p$ random points $P_0$ in the initial image. Given a depth image for the initial frame, we leverage a subset of the predicted tracks $\tau_{obj}$ (corresponding to moving points) to infer rigid-transforms of the object being manipulated $[\mathbf{T}_t]_{t=1}^H$ and show that these allow obtaining an open-loop plan in the form of robot end-effector poses $[\bar{\mathbf{a}}_t]_{t=1}^H$. Finally, we consider training a closed-loop residual policy $\pi_{\text{res}} (\mathbf{I}_t,\mathcal{G}, \mathbf{\tau},[\bar{\mathbf{a}}_t]_{t=1}^H)$ that corrects the open-loop action sequence $[\bar{\mathbf{a}}_t]_{t=1}^H$ by predicting residual actions at each timestep $\Delta\mathbf{a}_t$, such that the executed action sequence is $[\mathbf{a}_t = \bar{\mathbf{a}}_t + \Delta \mathbf{a}_t]_{t=1}^H$. In the subsequent sections, we explain the architecture and algorithm design for each of the three stages in our approach.

\begin{algorithm*}[t]
\caption{Predicting Rigid Transforms from Point Tracks}
\begin{algorithmic}[1]

\Procedure{Rigid Transforms}{$ \tau,\mathbf{I_0},\mathcal{G},P_0^{\text{3D}}, \mathbf{K}, H$}       
 \State $\{\{(x^i_t,y_t^i)\}_{i=1}^p\}_{t=1}^H = \mathbf{\tau}_{obj} = \texttt{filter}(\tau)$ \Comment{Filter moving point tracks}
    \State Unknown rigid transforms $[\mathbf{T}_t]_{t=1}^H$ \Comment{$\mathbf{T}_t$ has dimension 3x4}
    \State Run RANSAC on $\mathbf{\tau}_{obj}$ to filter outliers \Comment{optional}
    \For{$t \gets 1$ to $H$}  
        \State $\mathbf{T}_t = \argmin_{\mathbf{T}_t}\sum_i^N (||x_t^i-u_t^i|| + ||y_t^i-v_t^i||)$
        \State where $(u_t^i,v_t^i,1)  \simeq \mathbf{K}\mathbf{T}_tP_t$ \Comment{projections in homogeneous coordinates}
    \EndFor 
    \State \textbf{return} $\{\mathbf{T}_t=(\mathbf{R}_t,\mathbf{t}_t)\}_{t=1}^T$
\EndProcedure
\end{algorithmic}
\label{alg:rigid_transforms}
\end{algorithm*}
\vspace*{-0.2em}
\subsection{Point Track Prediction from Web Videos}
\vspace*{-0.2em}

We instantiate track prediction as a denoising process through a DiT based diffusion model~\cite{dit}. Let $\mathbf{I_0}$ denote the first frame of a video, and $\mathcal{G}$ denote the goal, which we consider to be the last frame of the video. For longer videos, we obtain multiple video clips of 4-5 seconds each for training. Let there be $p$ points in the initial frame to be tracked, such that $P_0$ denotes the set of those points and let $H$ be the prediction horizon. $[P_t]_{t=1}^H$ denotes the future locations of those points in the subsequent time-steps that we want to predict. In the forward diffusion process, all the points $P_t$ are corrupted by incrementally adding noise $\mathbf{\epsilon}_k$ ($k$ denotes the diffusion time-step), to obtain $\tilde{P}_t$, and converging to a unit Gaussian distribution $N(\mathbf{0},\mathbf{I})$. New samples can be generated by reversing the forward diffusion process, by going from Gaussian noise back to the space of point locations. To solve the reverse diffusion process, we need to train a noise predictor $\mathcal{V}_\theta(\mathbf{I_0},\mathcal{G},P_0, k)$. We design a DiT Transformer based architecture~\cite{dit} for $\mathcal{V}_\theta$ illustrated visually in Fig.~\ref{fig:arch}. Different from the original DiT model, we condition on embeddings of initial ($z_0$) and goal ($z_g$) images in addition to that of the diffusion step ($z_k$). The input to the Transformer in each batch is a sequence of $p$ tokens corresponding the number of points specified for tracking. The initial $P_0$ points are not noisy, as is the convention in training conditional diffusion models on time-series data. 

We train the prediction model with web videos by considering variable number of initial points $p$ that need to be tracked. For flexible modeling, the locations of the $p$ points are also randomized, such that at test-time any set of points in the initial image can be specified. We do not make any assumptions on objects to be tracked or camera motions in the videos, and do not curate the training videos in any way apart from ensuring they are of 4-5 second duration. If the goal image is such that multiple objects have moved from the initial scene, or the camera has moved, the track prediction model will predict different groups of motions for different objects and also predict motions of background points to account for camera motion. However, for robot experiments, we consider only a single object to be manipulated at a time, which is indeed the case with several diverse real-world tasks.

\begin{figure*}[t]
    \centering
    \includegraphics[width=\textwidth]{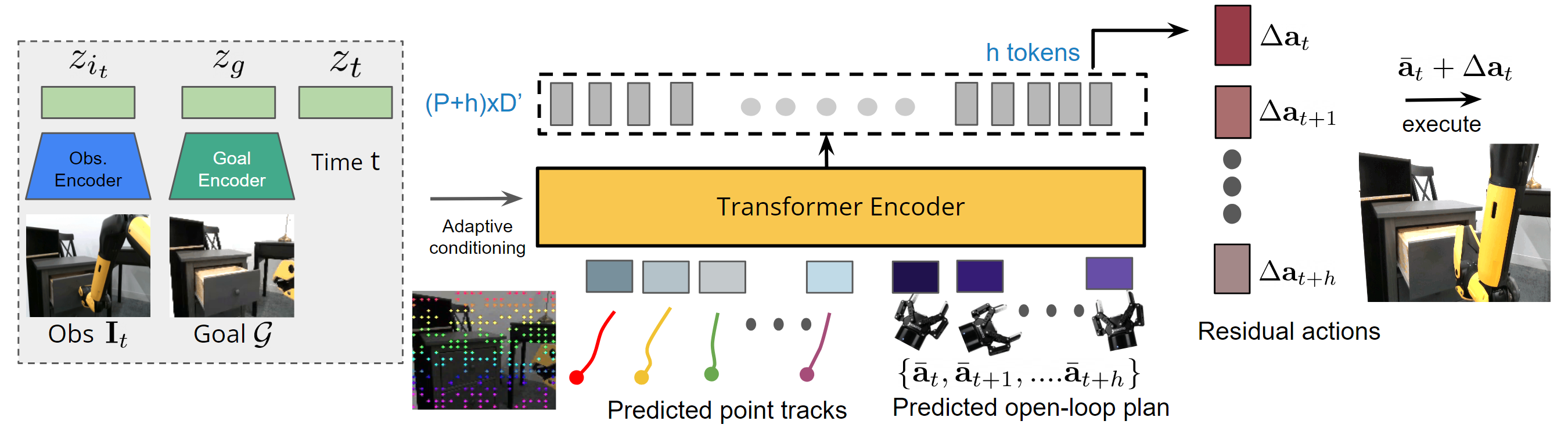}
    \caption{Architecture of the residual policy that predicts corrections at each time-step over the predicted open-loop plan, and enables closed-loop deployment.}
    \label{fig:policy_architecture}
    \vspace*{-2em}
\end{figure*}

\subsection{Inferring Coarse Manipulator Trajectory from Interaction Plan}
\label{sec:traj}

Given an image of a scene in a robot's environment $\mathbf{I}_0$, a goal $\mathcal{G}$, and a random set of points $P_0$ in the initial image, we can use the trained track prediction model to obtain future 2D locations of these points $\hat{P}_t$. As we consider scenarios with only a single object being manipulated under a fixed camera, only a subset of the points have a large predicted motion. We  identify these $p$ points and denote their predicted trajectories as  $\mathbf{\tau}_{\text{obj}} = [\{(x^i_t,y_t^i)\}_{i=1}^p]_{t=1}^H$. We consider the robot to be equipped with an RGBD camera, so we also have depth for the points $P_0$ in the first frame. Let us denote these 3D points as $P_0^{\text{3D}} = \{(x_0^i,y_0^i,z_0^i)\}_{i=1}^p$.


We seek to infer a (smooth trajectory of) per-time rigid transforms $\mathbf{T}_t$ of the object to be manipulated at time $t$ relative to first frame, given 3D points in the first frame $P_0^{\text{3D}}$, predicted 2D trajectory of points on the object $\mathbf{\tau}_{\text{obj}}$, and the camera intrinsic matrix $\mathbf{K}$. As described in Algorithm~\ref{alg:rigid_transforms}, these can be obtained by ensuring that the projection of the transformed 3D points i.e. $\mathbf{K}\mathbf{T}_t P_0$ matches the predicted 2D tracks $\{(x^i_t,y^i_t)\}_{i=1}^{p}$  as closely as possible at each time-step $t$. Let $\mathbf{K}\mathbf{T}_tP_0 \simeq \{(u_t^i,v_t^i,1)\}_{i=1}^{p}$. So the 2D projection of the $i^{\text{th}}$ point at time $t$ is $(u_t^i,v_t^i)$. Alternatively, we have the same coordinate for the point from the predicted track i.e. $(x^i_t,y_t^i)$. In order to determine the rigid transforms $\mathbf{T}_t$, we can solve the optimization problem in line 6 of Algorithm~\ref{alg:rigid_transforms} (e.g. with PnP solvers). This optimization is not under-constrained because the same 3D rigid transform $\mathbf{T}_i$ must explain the 2D motions of several points $P_0$ in the initial scene. Note that the obtained rigid transforms are \textit{embodiment-agnostic} and describe how the object should move in the scene.  

Now, to actually manipulate the object in the scene, we need to bring the robot end-effector\footnote{\footnotesize{by end-effector we mean the part of the robot that interacts with an object}} near the object, and optionally execute a grasp to hold on to the object, followed by transforming the end-effector based on the predicted rigid transforms  $[\mathbf{T}_t]_{t=1}^H$. For the first step, we use a heuristic such that given initial end-effector pose $\mathbf{e}_0$ we define the first transform $\mathbf{T}_0$ to be such that the end-effector moves to the center of the 3D points $\{(x^i_t,y^i_t)\}_{i=0}^{p}$ with the same orientation as $\mathbf{e}_0$. After moving the end-effector to this pose $\mathbf{e}_1$ we execute a grasp to hold the object. We obtain subsequent end-effector poses (open-loop action trajectory) by applying the rigid transforms $\mathbf{\bar{a}}_t = \mathbf{T}_t \mathbf{e}_1$.

\subsection{Closed-loop Manipulation with Residual Policy Correction}
\label{sec:residual}

The open-loop execution of the predicted 3D end-effector transforms described in the last section $[\bar{\mathbf{a}}_t]_{t=1}^H$ might fail due to small errors in the prediction. In addition, since the approach does not use any embodiment-specific data, it does not have accurate information for reasoning about contact with objects and might suffer from failures like being unable to grasp the object, in spite of executing the rest of the predicted trajectory correctly.  To remedy this, we propose learning a residual policy $\pi_{\text{res}} (\mathbf{I}_t,\mathcal{G}, \mathbf{\tau},[\bar{\mathbf{a}}_t]_{t=1}^H)$  shown in Fig.~\ref{fig:policy_architecture} to correct the predicted end-effector poses in each time-step. So the end-effector pose at time $t$ is 
\begin{equation}
    \hat{\mathbf{a}}_t = \Bar{\mathbf{a}}_t + \Delta\mathbf{a}_t \;\;;\;\;\;\text{where}\;\; \Delta\mathbf{a}_t = \pi_{\text{res}} (\mathbf{I}_t,\mathcal{G}, \mathbf{\tau},[\bar{\mathbf{a}}_t]_{t=1}^H)
\end{equation}
Instead of predicting just a single residual action $ \Delta\mathbf{a}_t$ we predict residuals $h$ steps in the future $ \Delta\mathbf{a}_{t:t+h}$ and during deployment execute just the first action. This multi-step prediction has been shown to mitigate compounding errors in behavior-cloning based training~\cite{act,bharadhwaj2023roboagent}. We can learn the residual policy with a small amount of robot demonstrations ($\sim 400$ trajectories overall) of representative tasks through behavior cloning. The data for each trajectory consists of observation-action pairs of the form $[(\mathbf{I}_t,\mathbf{a}_t)]_{t=1}^H$. Here, $\mathbf{I}_t$ denotes images observed from the robot's camera and $\mathbf{a}_t$ denotes actions in the form of end-effector poses. 

Crucially, since the aim of this policy is to learn only small corrections to the predicted waypoints $[\bar{\mathbf{a}}_t]_{t=1}^H$, we do not need to learn this policy with data from the exact scenarios that the system will be deployed in and the prediction model is expected to generalize to unseen scenarios by virtue of diverse training. The rationale is that having some embodiment-specific demonstration data in a few scenarios will help correct for the open-loop predictions from web-only data. For evaluation, we consider different levels of generalization with unseen object instances and completely unseen objects in unseen scenes.

\vspace*{-0.5em}
\section{Experiment Setup}
\vspace*{-0.5em}
We focus our experiments on in-the-wild manipulation scenarios where a mobile manipulator needs to manipulate objects in different living rooms, offices, and kitchens based on specified goals. For all the robot experiments, we use a Boston Dynamics Spot robot equipped with a manipulator (hand) and a front facing Intel RealSense camera~\cite{keselman2017intel}. We manipulate the arm through end-effector control based on the outputs of our policy.

\vspace*{-0.5em}
\subsection{Evaluation Details}
\label{sec:evaldetails}

\noindent{\textbf{Track Prediction}.} For quantitative evaluation of the track prediction model, we adopt a modification of the metric developed by prior works~\cite{cotracker,tapir}, $\delta^x_t$. For evaluation videos, we consider the output of Co-Tracker~\cite{cotracker} to be the ground-truth and compare the difference with respect to the predictions, based on the $\delta^x_t$ metric. We define $\delta^x_t$ to be the fraction of points that are within a threshold pixel distance of $x$ of their ground truth in a time-step $t$. We report the \textit{area under the curve} $\Delta$ with $\delta^x_t$ by varying $x$ from $1$ to $N=10$ and taking the average across the prediction horizon $H$ i.e. $\Delta = (\sum^H_{t=1}\sum_{x=1}^{N}\delta^x_t)/H$. Hence, $\Delta$ can vary from $0$ to $1$ with higher being better.

\noindent{\textbf{Track Prediction}.} As is the convention in goal-conditioned robot learning, we perform evaluations by quantifying success rate, where a successful trajectory is defined to be one where the final pose of the object in the scene to be manipulated is identical to the pose of the object in the goal image.  We categorize results based on different levels of generalization, the definitions of which are inspired by prior works~\cite{bharadhwaj2023roboagent,bharadhwaj2023towards,rt1,rt2}:

\begin{itemize}
    \item Mild Generalization (\textbf{MG}): unseen configurations of seen object instances in seen scenes; organic scene variations like lighting and background changes
    \item Standard Generalization (\textbf{G}): unseen object instances in seen/unseen scenes
    \item Combinatorial Generalization (\textbf{CG}): unseen activity-object type combinations in seen/unseen scenes
    \item Type Generalization (\textbf{TG}): completely unseen object types, or completely unseen activities, in unseen scenes
\end{itemize}

\subsection{\textbf{Baselines and Comparisons}}

For quantitative evaluations, we first compare our track prediction approach with other related baselines and then perform comparisons with baselines for robot manipulation experiments.

\noindent{\textbf{Track Prediction}.} We perform comparisons with two baselines that have the same inputs as our track prediction model, i.e. an initial image, a goal image and points specified on the initial image, and the same output type i.e. point tracks in between the initial and goal images. We compare with a flow-based baseline that directly predicts flow between the initial and goal images, and then performs a per-timestep interpolation of the flow vectors~\cite{xu2023unifying}. The second baseline performs video-infilling given initial and goal images~\cite{fu2023tell}, and then uses Co-Tracker~\cite{cotracker} to obtain tracks on the generated video.

\noindent{\textbf{Robot Experiments}.} We perform several comparisons with baselines and ablation studies for goal-conditioned robot manipulation. For baselines, we use the same embodiment-specific demonstrations as \textit{Ours}, the goal-conditioned policy that predicts residuals over open-loop actions at each time-step.
\vspace*{-0.2em}
\begin{itemize}
    \item \textit{Goal-Conditioned BC} is a baseline for multi-task policy learning, similar to prior works~\cite{rt1,rt2,bharadhwaj2023roboagent}. 
    \item \textit{Affordance-Conditioned BC} is the approach from~\cite{bahl2023affordances} that conditions the policy on predicted affordances in the initial image.
    \item  \textit{Video-Conditioned BC} based on~\cite{fu2023tell,unipy,ko2023learning} first predicts RGB video and then does tracking on top of it.
    \item \textit{Hand-Object Mask Conditioned BC} from~\cite{bharadhwaj2023towards} conditions the policy on a predicted interaction plan consisting of masks of hands and objects. 
\end{itemize}
\vspace*{-0.2em}
\noindent\textit{Ours (Open Loop)} is the approach for track prediction followed by open-loop execution as described in Algorithm~\ref{alg:rigid_transforms}. This does not use any embodiment-specific data for training. To understand the benefit of predicting residuals over actions as opposed to predicting complete actions, we compare with an ablated variant \textit{ Ours (actions; not residuals)} that predicts actions $\hat{\mathbf{a}}_t$ directly without predicting residuals $\Delta\mathbf{a}_t$ and not relying on an open-loop plan as input.

\vspace*{-0.2em}
\subsection{Training Data}

For training the track prediction model, we leverage diverse passive videos available on the web that are not collected by us. Specifically, we use human video clips from EpicKitchens~\cite{epic} (clipping videos to ensure they are of 4-5 seconds duration), and large-scale robot videos released in RT1 data~\cite{rt1} and BridgeData~\cite{walke2023bridgedata}. To obtain ground-truth tracks for training the prediction model, we run Co-Tracker~\cite{cotracker} on the resulting 400,000 video clips. Note that the robot datasets (RT1 and Bridge) are on completely different robots and scenarios than the robot we use for experiments (Spot). For training the residual policy, the embodiment-specific data we collect consists of $\sim 400$ trajectories obtained by tele-operating the Spot, for solving 10 tasks of manipulating everyday objects like doors, drawers, bottles, jugs. Note that this embodiment-specific data we collect is 3-4 orders of magnitude less than that what related works~\cite{rt1,rt2,bharadhwaj2023roboagent} require for policy learning. 

\section{Results}
\vspace*{-0.5em}

We present qualitative results of the predicted tracks, and robot evaluations, followed by quantitative comparisons with the metrics defined in section~\ref{sec:evaldetails}. Please refer to the supplementary zip for detailed qualitative results and robot evaluation videos.

\vspace*{-0.5em}
\subsection{Point Track Prediction Results}
    \vspace*{-1em}
\begin{figure*}[t]
    \centering
    \includegraphics[width=0.95\textwidth]{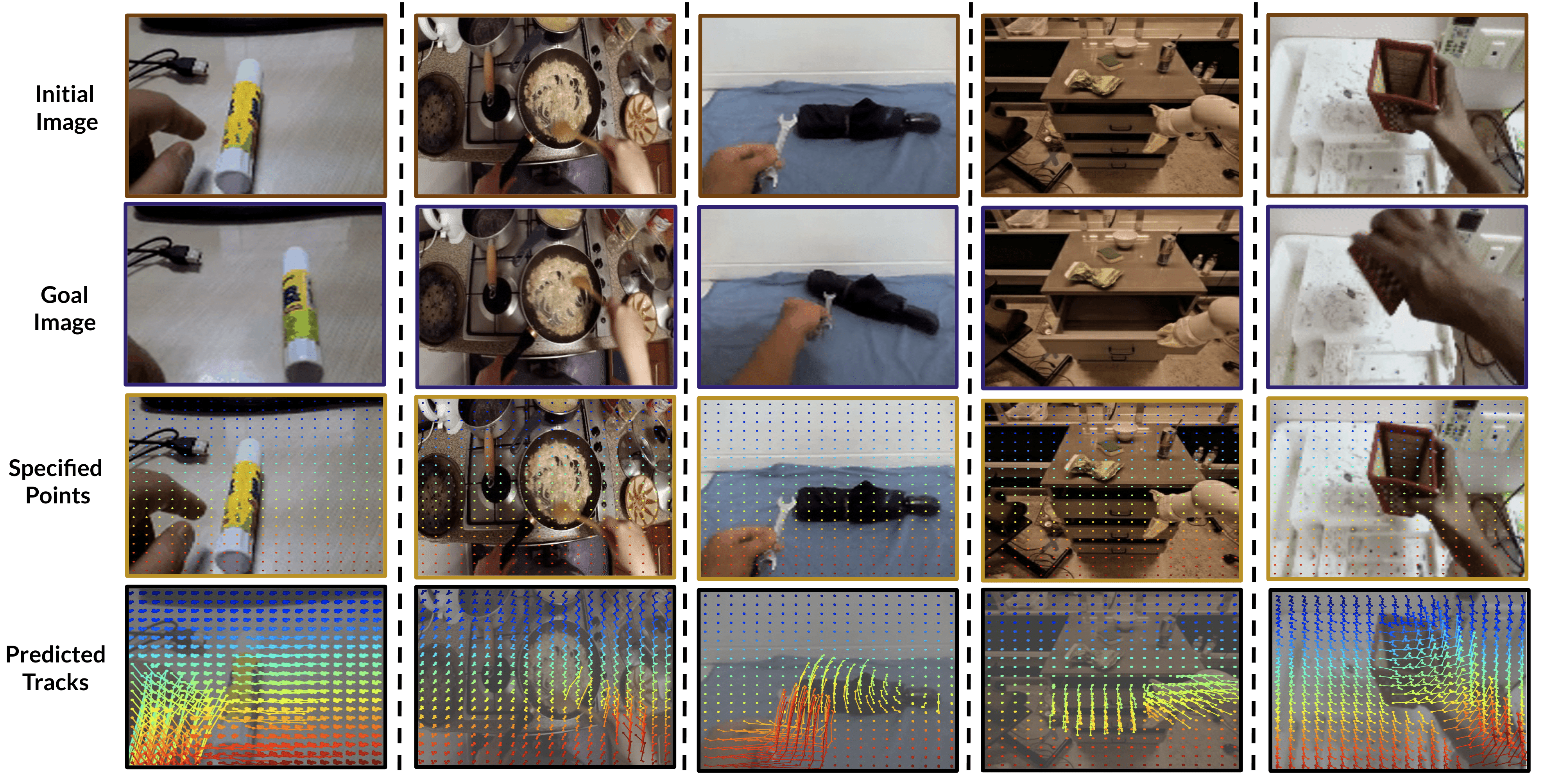}
    \caption{We show qualitative results of the track predictions on unseen initial and goal images across diverse datasets. Given specified points on the initial image we predict future tracks of these points, corresponding to the goal image. We can see that the predictions are plausible and correspond to manipulating the object(s) in the scene. }
    \label{fig:qual_predictions}
    \vspace*{-1.5em}
\end{figure*}

\setlength{\tabcolsep}{3pt}
\begin{table}[h!]
\centering
\caption{Evaluation of track prediction performance on held-out videos from different datasets on the web. EpicKitchens~\cite{epic} and SmthSmthv2~\cite{smthsmth} are datsets of human videos, and BridgeData~\cite{walke2023bridgedata} and RT1 data~\cite{rt1} are datasets of robot videos. Note that we train a \textit{single} model that we evaluate on these different datasets. The metric $\Delta$ is defined in section~\ref{sec:evaldetails}. Higher is better and the range is from 0 to 1.}
\begin{tabular}{@{}ccccc@{}}
\toprule
                  & \textbf{EpicKitchens}~\cite{epic} & \textbf{SmthSmthv2}~\cite{smthsmth} & \textbf{BridgeData}~\cite{walke2023bridgedata} & \textbf{RT1 Data}~\cite{rt1} \\ \midrule
\textbf{Flow}~\cite{xu2023unifying}     &           0.21            &     \cellcolor{yellow!25}  0.27               &          0.42           &      0.38             \\ 
 \textbf{Video}~\cite{fu2023tell}     &         \cellcolor{yellow!25}   0.30            &          0.17           &          -           &           -        \\              \textbf{Ours} &  \cellcolor{yellow!75} 0.67                    &  \cellcolor{yellow!75} 0.70                  &  \cellcolor{yellow!75} 0.77                  &  \cellcolor{yellow!75} 0.75                \\ \bottomrule
\end{tabular}
\label{tb:pred}
\vspace*{-1em}
\end{table}

We first look at some qualitative results of the track prediction model in different unseen scenes. In Fig.~\ref{fig:qual_predictions} we show visualization of track predictions on unseen initial and goal images across diverse datasets. We choose points on a grid in the initial frame, as shown in the third row. The prediction model is conditioned on the initial image, the goal image, and the set of points in the initial image whose future tracks are to be predicted. We can see that the predictions (shown in the fourth row) are plausible and correspond to manipulating the objects in the scene as described by the respective goal images. We can also see that when multiple entities (e.g. human and object or robot and object) or the camera moves between the initial and goal images, there are different sets of point tracks predicting the respective motions.

In Table~\ref{tb:pred} we perform evaluations for track prediction by comparing with the flow-based~\cite{xu2023unifying} and video-based~\cite{fu2023tell} baselines. We can see that both the baselines have much lower accuracy compared to our approach of predicting point tracks. This is because flow is too coarse to capture large non-linear state changes in between the initial and goal images. Whereas, predicting an RGB video followed by tracking suffers due to issues of implausible generation because video generation is a much more complex task than predicting the tracks of a set of points where the details about appearance, texture etc. are abstracted out. For reference, not predicting any movement for any point at all time-steps scores 0.03, 0.05, 0.36, 0.28. This suggests the benefit of directly predicting future point tracks as done by our approach if the aim is to capture motion of objects in the scene between an initial and a goal image.

\begin{figure*}[t]
    \centering
    \includegraphics[width=1.0\textwidth]{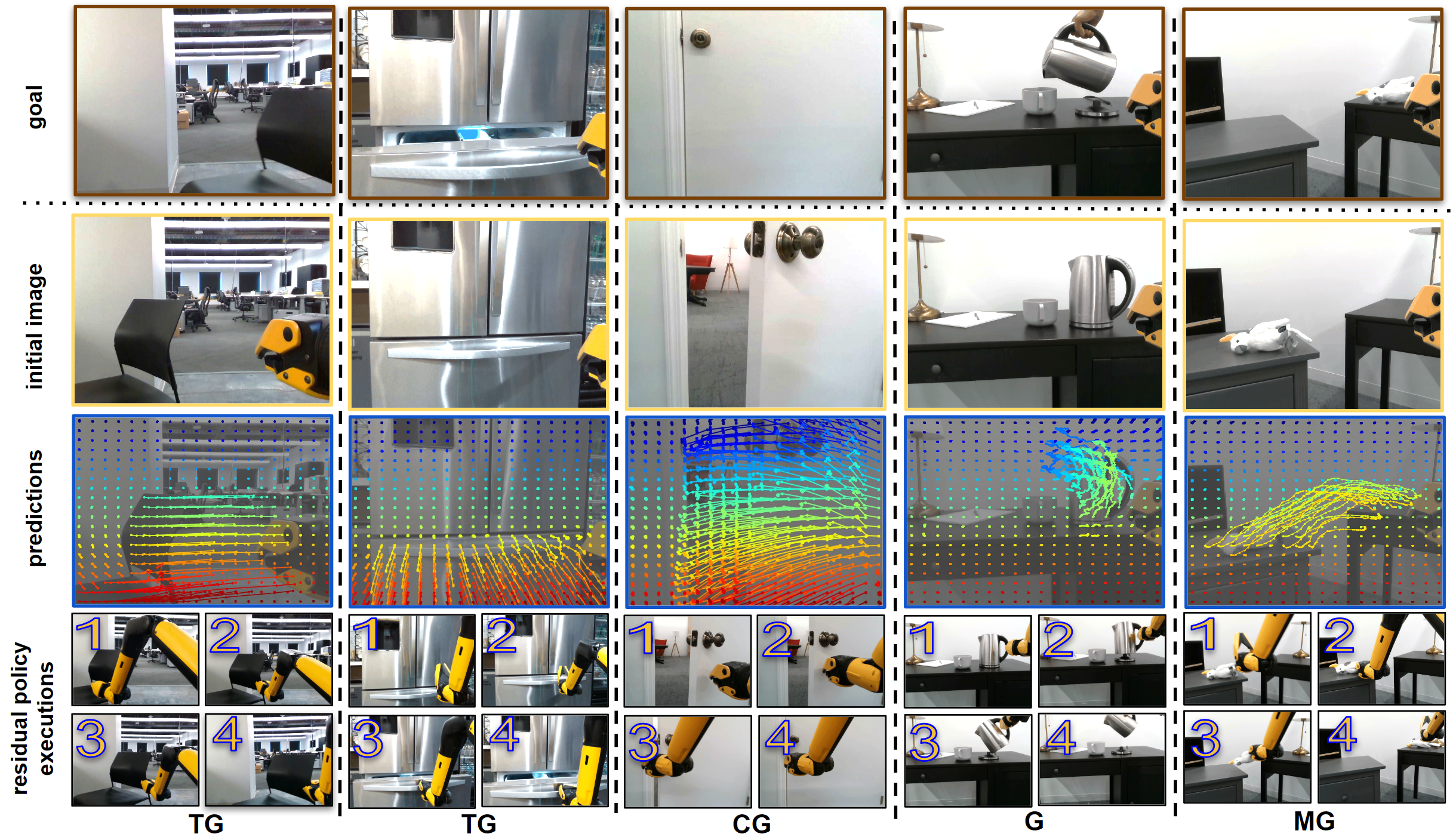}
    \caption{We show visualizations of point track predictions for different tasks, followed by closed-loop execution with the residual policy. We can see that the predictions are plausible and the robot execution successfully realizes the predictions to complete the respective tasks specified by the goal images. The bottom row shows the generalization level for each task, defined in section~\ref{sec:evaldetails}.}
    \label{fig:qual_executions}
    \vspace*{-2em}
\end{figure*}
\subsection{Robot Manipulation Results} 

\begin{figure*}[t]
    \centering
    \includegraphics[width=0.95\textwidth]{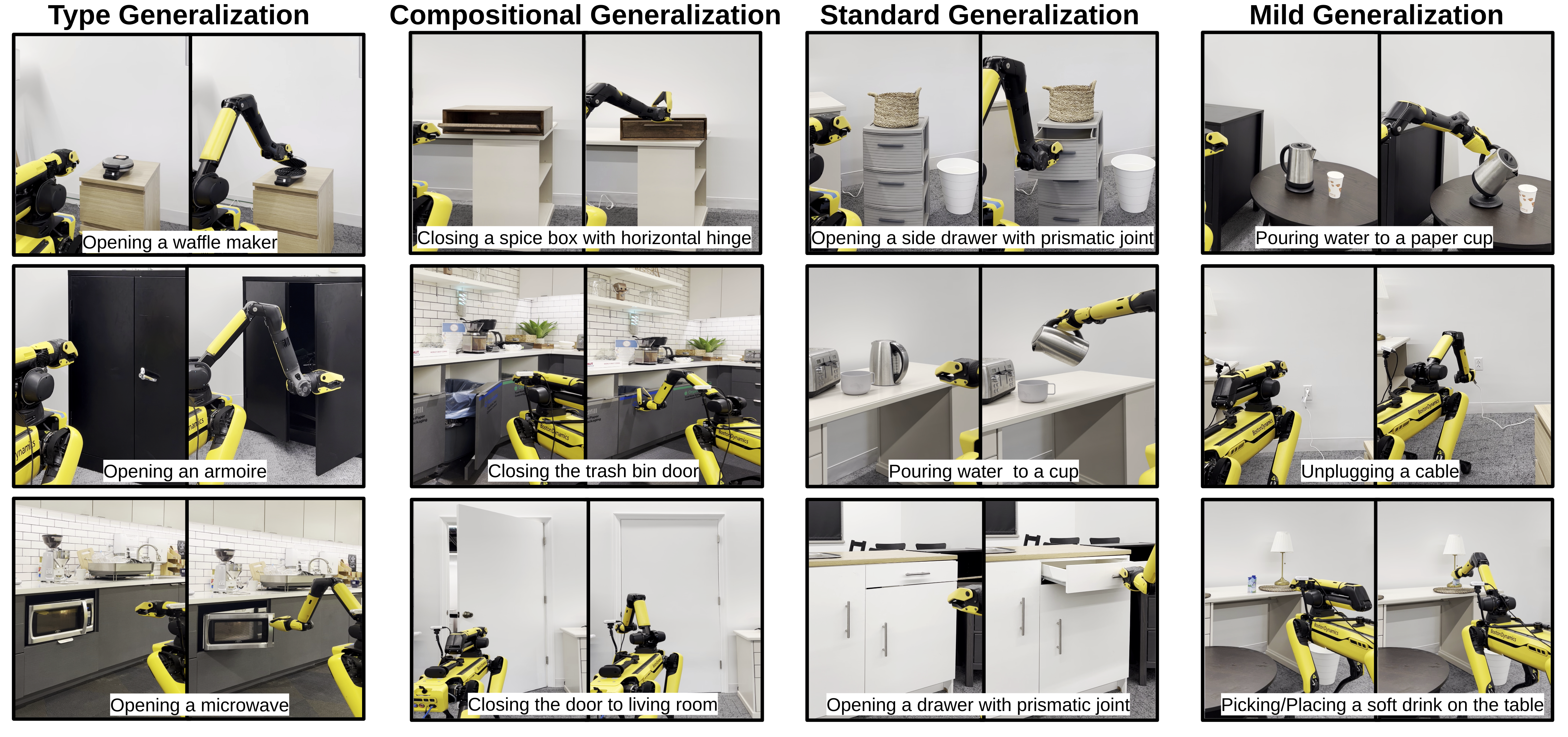}
    \caption{Qualitative results showing robot executions (from a third person view) with the residual policy for different tasks with respect to the generalization levels defined in section~\ref{sec:evaldetails}. We show the first and last images of a rollout. The robot executions are best viewed as videos in the supplementary zip.}
    \label{fig:qual_robot}
        \vspace*{-2em}
\end{figure*}

We visualize results of point track predictions using the trained prediction model in a robot's environment, followed by residual policy executions based on the predictions.
In Fig.~\ref{fig:qual_executions} we show the track predictions overlayed on the initial image, corresponding to the goal image shown in the top row. The bottom row shows robot execution with the first frame (1), two intermediate frames (2,3) and the last frame (4) of a rollout. We can see that the predicted tracks correspond to points on the object moving in a way that satisfies the goal, and the policy is able to manipulate the object to the desired goal configuration. Since the camera doesn't move between the initial and goal images, we can see that background (non object) pixels remain stationary in the predictions, which is useful for accurate prediction of rigid transforms of the object.

In Table~\ref{tb:robot} we show comparisons for robot manipulation experiments, respectively for each level of generalization. We evaluate each approach for 20 rollouts in each level, across a total of 25 tasks in 5 different physical kitchen, office, and living room locations.  We first note that our residual policy outperforms our approach for directly executing an open-loop plan based on predicted rigid transforms. This suggests that the residual policy is able to correct for inaccuracies in the open-loop plan by virtue of leveraging some embodiment-specific data that helps in performing accurate grasps on objects and recovering from potential failures during a trajectory.

We observe that for mild generalization (MG), the goal-conditioned BC baseline has slightly lower success rate compared to our residual policy, and significantly lower (or zero) success rates for standard (G), compositional (CG), and type (TG) generalization. This suggests the benefit of leveraging web video data for learning interaction plans that helps our approach generalize effectively. Finally, compared to baselines that also leverage web data like affordance-conditioned BC, video-conditioned BC, and hand-object mask-conditioned BC, we observe significant gains from our approach in the higher levels of generalization (CG and TG). This suggests that predicting static affordances without reasoning about motion trajectories, hallucinating RGB videos that suffer from incorrect generations and produce implausible artifacts in the scene, or predicting 2D masks of hands and objects without reasoning about correspondences are insufficient cues for effectively leveraging web videos. Compared to these, our interaction plan learned through track prediction provides sufficient cues for solving unseen manipulation tasks by virtue of allowing inference of 3D rigid transforms, and the residual policy helps correct for errors in the predictions.   


\setlength{\tabcolsep}{8pt}
\begin{table}[t]
\centering
\caption{Evaluation of goal-conditioned robot manipulation experiments, per the protocol described in section~\ref{sec:evaldetails}. The numbers denote success rate averaged over 20 rollouts for different tasks within each generalization axis (Higher is better). Detailed list of tasks are in the Supplementary pdf. Refer to Fig.~\ref{fig:qual_robot} for visualizations of some task rollouts corresponding to each of the four generalization axes.}
\begin{tabular}{@{}ccccc@{}}
\toprule
                                    & \textbf{MG} & \textbf{G} & \textbf{CG} & \textbf{TG} \\ \midrule
\textbf{Behavior Cloning (BC)}       &        60\%                           &         20\%                             &         0\%                            &                   0\%                     \\ 
\textbf{Affordance-Conditioned}       &                65\%                   &      30\%                                &                  10\%                   &     5\%                                   \\ 
\textbf{Video-Conditioned}       &                60\%                   &      25\%                                &                  0\%                   &     0\%                                   \\ 
\textbf{Hand-Object Mask-Conditioned}       & \cellcolor{yellow!75}  70\%                                &           40\%                           &         25\%                            &            20\%                           \\ \hline
\textbf{Ours (Open-Loop)} &   35\%                                &     25\%                                 &  \cellcolor{yellow!25}  30\%                                 &       25\%                                 \\
\textbf{Ours (Ablation; actions not residuals)}       &    \cellcolor{yellow!25}        70\%                       &   \cellcolor{yellow!75}     45\%                              &     \cellcolor{yellow!25}          30\%                      &    \cellcolor{yellow!25}   30\%       \\ 
\textbf{Ours}     &   \cellcolor{yellow!75}70\%                                &       \cellcolor{yellow!75} 60\%                               &     \cellcolor{yellow!75}   55\%                              &  \cellcolor{yellow!75}      40\%                                 \\
\bottomrule
\end{tabular}
\label{tb:robot}
\vspace*{-2em}
\end{table}
\subsection{Analysis of Failures}
Here we discuss the failures displayed by our framework. For the open-loop plan based on residual transforms, the main failure modes we observe are inability to grasp the object at the right location, and inability to recover from intermediate failures. The residual policy corrects for these behaviors by virtue of leveraging some embodiment-specific data, and thus has higher success rates. We note that the success rate for higher levels of generalization CG and TG) is still not very high since these are very challenging settings and the residual policy sometimes fails by incorrectly grasping the object, getting stuck during the execution by trying to execute a non-feasible motion, or by executing a trajectory that does not conform with the goal image specified.

\vspace*{-0.5em}
\section{Discussion and Conclusion}
\vspace*{-0.5em}
In this paper, we developed a framework for generalizable robot manipulation by leveraging large-scale web video data to learn embodiment agnostic plans of how objects should be manipulated in a scene to satisfy a goal. We combined this with a small amount of embodiment-specific data to learn residual corrections over the predicted plans through a closed-loop policy. Our real world manipulation results across a range of diverse tasks with varying levels of generalization demonstrate the potential of scalably leveraging web data to predict plans for object manipulation. While our framework allows for strong generalization to unseen tasks in-the-wild, the tasks are still of short-horizon and involve manipulating a single object in the scene. It would be an interesting direction of future work to extend our framework for tackling long-horizon tasks that involve successive manipulations of multiple objects in the scene.  
\section*{Acknowledgement}
We thank Yufei Ye, Himangi Mittal, Devendra Chaplot, Abitha Thankaraj, Tarasha Khurana, Akash Sharma, Sally Chen, Jay Vakil, Chen Bao, Unnat Jain, Swaminathan Gurumurthy for helpful discussions and feedback.  We thank Carl Doersch and Nikita Karaev for insightful discussions about point tracking. This research was partially supported by a Google gift award.

%
%
\bibliographystyle{splncs04}
\bibliography{references}
\newpage
\clearpage
\section*{Appendix}

\subsection{Video results}
Please refer to the supplementary website \url{https://homangab.github.io/track2act/} for detailed qualitative results of our framework including robot video evaluations.

\subsection{Robot Experiment Details}
We perform all robot manipulation experiments with a Boston Dynamics Spot Robot, operated through end-effector control. The robot is a quadruped with an arm attached to its base. We connect a front-facing Intel Realsense camera to the base such that it always moves with the robot, and it static with respect to the base. The end-effector of the arm is a two-fingered gripper. The horizon $H$ of rollouts is 50 steps, and we operate the robot at a frequency of 5 Hz. For the residual policy, at each step we predict actions $h=4$ time-steps in the future, and execute the first action. We execute the predicted actions on the robot through an Inverse Kinematics (IK) controller. This controller converts the end-effector poses to robot joint actions for appropriately manipulating the arm. We use the IK controller provided by Boston Dynamics for this purpose.

\subsection{Track Prediction Model details}
We instantiate track prediction as a denoising process through a DiT based diffusion model~\cite{dit}. Let $\mathbf{I_0}$ denote the first frame of a video, and $\mathcal{G}$ denote the goal, which we consider to be the last frame of the video. For longer videos, we obtain multiple video clips of 4-5 seconds each for training. Let there be $p$ points in the initial frame to be tracked, such that $P_0$ denotes the set of those points and let $H$ be the prediction horizon. $[P_t]_{t=1}^H$ denotes the future locations of those points in the subsequent time-steps that we want to predict. In the forward diffusion process, all the points $P_t$ are corrupted by incrementally adding noise $\mathbf{\epsilon}_k$ ($k$ denotes the diffusion time-step), to obtain $\tilde{P}_t$, and converging to a unit Gaussian distribution $N(\mathbf{0},\mathbf{I})$. New samples can be generated by reversing the forward diffusion process, by going from Gaussian noise back to the space of point locations. To solve the reverse diffusion process, we need to train a noise predictor $\mathcal{V}_\theta(\mathbf{I_0},\mathcal{G},P_0, k)$. We design a DiT Transformer based architecture~\cite{dit} for $\mathcal{V}_\theta$ illustrated visually in Fig.~\ref{fig:arch}. Different from the original DiT model, we condition on embeddings of initial ($z_0$) and goal ($z_g$) images in addition to that of the diffusion step ($z_k$). The input to the Transformer in each batch is a sequence of $p$ tokens corresponding the number of points specified for tracking. The initial $P_0$ points are not noisy, as is the convention in training conditional diffusion models on time-series data. We train the prediction model with web videos by considering variable number of initial points $p$ that need to be tracked. We vary $p$ from 200 to 400. For flexible modeling, the locations of the $p$ points are also randomized, such that at test-time any set of points in the initial image can be specified. We do not make any assumptions on objects to be tracked or camera motions in the videos, and do not curate the training videos in any way apart from ensuring they are of 4-5 second duration. 

The model has 24 DiT blocks, with a hidden size of 1024, and 16 heads. The ResNet18 embeddings of initial image and goal image have dimensions 512. The condition to each DiT block consists of the sum of initial image embedding, goal image embedding, and diffusion time-step embedding through adaptive modulation (adaLN) layers. The adaptive modulation layers and final MLP layers are zero-initialized, and the rest are Xavier uniform initialized. We use Adam optimizer with default Adam betas = (0.9,0.999) and a constant learning rate of 1e-4 for experiments. The rest of the architecture and training details are similar to DiT~\cite{dit}.

\subsection{Residual Policy Model details}
 To correct the predicted open-loop plan, with a small amount of embodimen-specific data, we propose learning a residual policy $\pi_{\text{res}} (\mathbf{I}_t,\mathcal{G}, \mathbf{\tau},[\bar{\mathbf{a}}_t]_{t=1}^H)$  shown in Fig.~\ref{fig:policy_architecture} to correct the predicted end-effector poses in each time-step. So the end-effector pose at time $t$ is 
$
    \hat{\mathbf{a}}_t = \Bar{\mathbf{a}}_t + \Delta\mathbf{a}_t \;\;;\;\;\;\text{where}\;\; \Delta\mathbf{a}_t = \pi_{\text{res}} (\mathbf{I}_t,\mathcal{G}, \mathbf{\tau},[\bar{\mathbf{a}}_t]_{t=1}^H) $
Instead of predicting just a single residual action $ \Delta\mathbf{a}_t$ we predict residuals $h$ steps in the future $ \Delta\mathbf{a}_{t:t+h}$ and during deployment execute just the first action. This multi-step prediction has been shown to mitigate compounding errors in behavior-cloning based training~\cite{act,bharadhwaj2023roboagent}. We can learn the residual policy with a small amount of robot demonstrations ($\sim 400$ trajectories overall) of representative tasks through behavior cloning. The data for each trajectory consists of observation-action pairs of the form $[(\mathbf{I}_t,\mathbf{a}_t)]_{t=1}^H$. Here, $\mathbf{I}_t$ denotes images observed from the robot's camera and $\mathbf{a}_t$ denotes actions in the form of end-effector poses. 

The residual policy model is a Transformer based on the DiT architecture. The model has 12 DiT blocks, with a hidden size of 512, and 8 heads. The ResNet18 embeddings of initial image and goal image have dimensions 512. The condition to each DiT block consists of the sum of current image embedding, goal image embedding, and emebedding of the current time-step $t$  through adaptive modulation (adaLN) layers. The adaptive modulation layers and final MLP layers are zero-initialized, and the rest are Xavier uniform initialized. We use Adam optimizer with default Adam betas = (0.9,0.999) and a constant learning rate of 1e-4 for experiments. The input to the model consists of the predicted tracks of $p$ points in the initial image (we keep $p=400$ to ensure a dense grid in the initial image of dimensions 256x256x3) and the predicted open-loop plan with $h$  steps from $t:t+h$. So there are $p+h$ input tokens. We read off the final $h$ tokens corresponding to the updated open-loop plan for these $h$ steps and after a final MLP layer, output actions for $h$ steps .  We will release all code and models upon acceptance.

\subsection{Training Data for Track Prediction}
We use four different web data sources for training the track prediction model - videos from Something-Something-v2~\cite{smthsmth}, Epic-Kitchens~\cite{epic}, RT1 data~\cite{rt1}, and BridgeData~\cite{walke2023bridgedata}. Something-Something-v2 contains short YouTube videos of people doing everyday activities. We consider videos from this dataset as is, and choose the first frame as the initial image and the last frame as goal image. Epic-Kitchens contains ego-centric videos of humans in different locations performing diverse tasks in kitchens. Since these videos are long ($\geq$ 20 min each), we choose clips of duration 4-5 seconds by cutting the long videos, and choosing clips where a human hand is visible in the scene (so as to have clips where an object is being manipulated, instead of a person just moving around). RT1 Data and Bridge Data are large-scale robot datasets that contains rollouts of two different types of robots being tele-operated for different tasks. For these datasets, we consider the first and last images to be the first and last frames of a rollout, and each rollout to be a separate video.

In total we obtain around 400,000 videos clips from the above sources, choose a dense grid of 400 points on the first frame and we run Co-Tracker~\cite{cotracker} on these clips, for obtaining the ground-truth intermediate tracks of points. Our prediction model is conditioned on the first and last frames for each video, and the task of predicting the tracks of random points on the initial frame is supervised by the tracks we obtain from Co-Tracker (ground-truth). 

\subsection{Training Data for Residual Policy}
For training the residual policy we collected tele-operated demonstrations with the Spot robot by controlling it with a joystick across 10 tasks in 3 physical locations. These scenarios correspond to only a subset of the diverse tasks, objects, and scenes we consider for evaluation . Concretely, the evaluation scenarios with same tasks as the collected data correspond to the \textit{mild generalization} (MG) category. Rest of the generalization axes corresponding to unseen instances and categories are described in detail in section~\ref{sec:evaldetails}. 

The training data consists of 400 teleoperated trajectories, each consisting of $H$ (observation,action) pairs ($H=50$).  The data for each trajectory consists of observation-action pairs of the form $[(\mathbf{I}_t,\mathbf{a}_t)]_{t=1}^H$. Here, $\mathbf{I}_t$ denotes images observed from the robot's camera and $\mathbf{a}_t$ denotes actions in the form of end-effector poses. This data is collected at the same frequency of 5 Hz that we deploy the policy for eventual evaluations. Note that this embodiment-specific data we collect is 3-4 orders of magnitude less than that what related works~\cite{rt1,rt2,bharadhwaj2023roboagent} require for policy learning. This is a major advantage of our framework as it precludes the need to spend years on real-world data collection, while achieving generalization to more diverse scenarios by virtue of leveraging \textit{passive} web videos for track prediction.

\subsection{Details on baselines}

We perform several comparisons with baselines and ablation studies for goal-conditioned robot manipulation. For baselines, we use the same embodiment-specific demonstrations as \textit{Ours}, the goal-conditioned policy that predicts residuals over open-loop actions at each time-step (Algorithm~\ref{alg:residual}).
\vspace*{-0.2em}
\begin{itemize}
    \item \textit{Goal-Conditioned BC} is a baseline for multi-task policy learning, similar to prior works~\cite{rt1,rt2,bharadhwaj2023roboagent}. This is trained with the same data we use for training our residual policy, and is conditioned on goal image, similar to our residual policy.
    \item \textit{Affordance-Conditioned BC} is the approach from~\cite{bahl2023affordances} that conditions the policy on predicted affordances in the initial image. These affordances capture what is \textit{plausible} to be manipulated in the scene, and so are different from our time-series predictions of point tracks. We directly adopt the affordance model from~\cite{bahl2023affordances} that was trained on web data, and use the same embodiment-specific data as our residual policy for training through conditional behavior cloning.  
    \item  \textit{Video-Conditioned BC} based on~\cite{fu2023tell,unipy,ko2023learning} first predicts RGB video and then does tracking on top of it. We adopt the video prediction model from~\cite{fu2023tell} (without language conditioning) trained on web data, and use the same embodiment-specific data as our residual policy for training through conditional behavior cloning.   
    \item \textit{Hand-Object Mask Conditioned BC} from~\cite{bharadhwaj2023towards} conditions the policy on a predicted interaction plan consisting of masks of hands and objects. We use the hand-object plan prediction model from~\cite{bharadhwaj2023towards}, and use the same embodiment-specific data as our residual policy for training through conditional behavior cloning. Note that this baseline is slightly different from the translation model in~\cite{bharadhwaj2023towards} because we do not collect paired human-robot demonstrations unlike~\cite{bharadhwaj2023towards} and so the policy is conditioned on predicted hand-object plans as opposed to ground-truth plans unlike~\cite{bharadhwaj2023towards}. 
\end{itemize}
\vspace*{-0.2em}

Comparison to \textit{Goal-Conditioned BC} helps understand the potential benefits of leveraging web data for generalizable manipulation, and comparisons to \textit{Affordance-Conditioned BC}, \textit{Video-Conditioned BC}, \textit{Hand-Object Mask Conditioned BC} help understand the potential of predicting point tracks from web videos, compared to other ways of using web data for prediction geared towards manipulation.

\noindent\textit{Ours (Open Loop)} is the approach for track prediction followed by open-loop execution as described in Algorithm~\ref{alg:rigid_transforms}. This does not use any embodiment-specific data for training. To understand the benefit of predicting residuals over actions as opposed to predicting complete actions, we compare with an ablated variant \textit{ Ours (actions; not residuals)} that predicts actions $\hat{\mathbf{a}}_t$ directly without predicting residuals $\Delta\mathbf{a}_t$ and not relying on an open-loop plan as input.

\subsection{Qualitative Results for baselines}
We provide qualitative comparisons of the baselines with our approach, in the figures below. For detailed qualitative video results of our approach, please refer to the website.

\begin{figure}
    \centering
   \includegraphics[width=\linewidth]{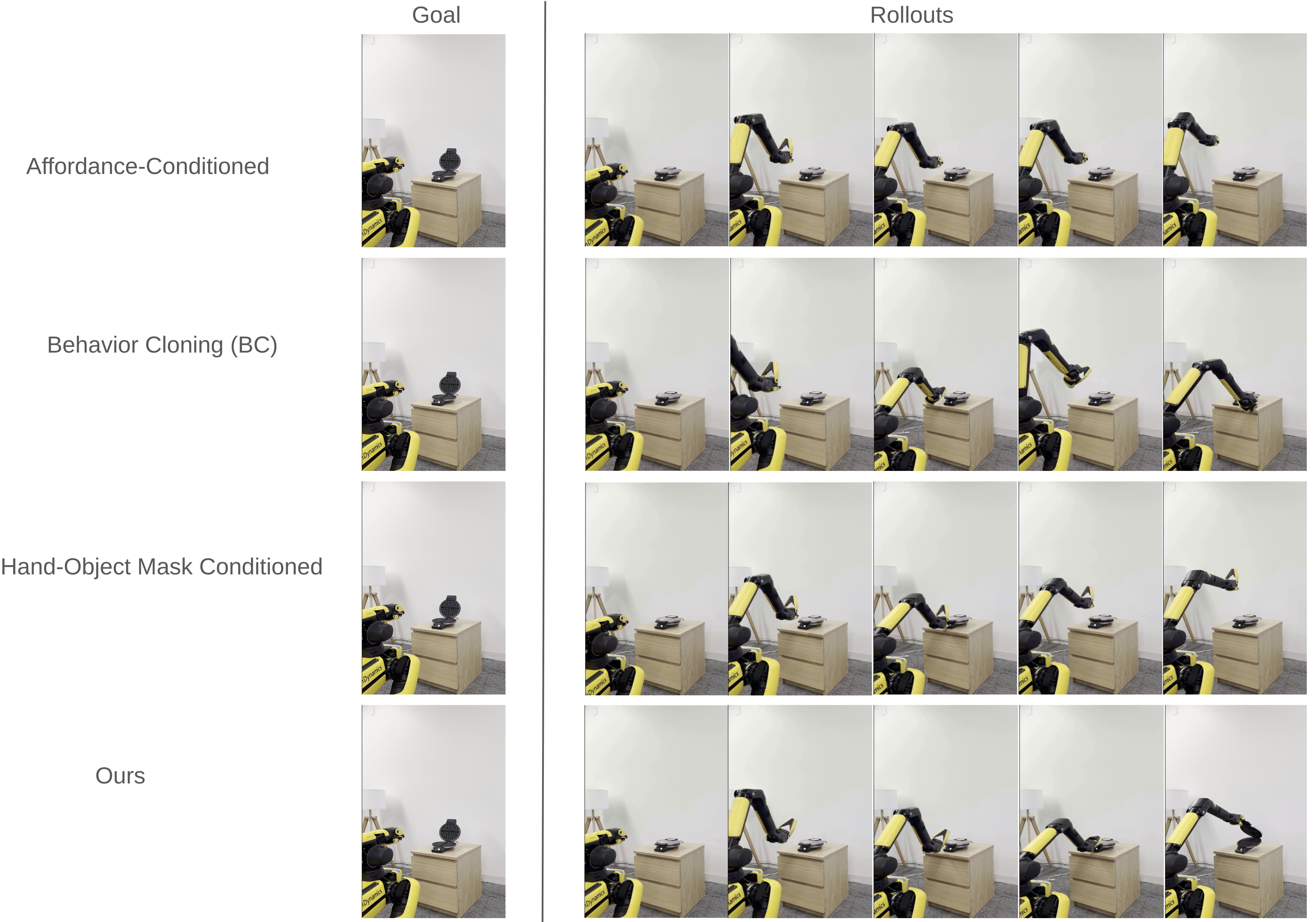}
    \caption{Type Generalization (TG). We show rollouts from baselines for the same goal. The views are from a third person camera.}
    \label{fig:enter-label}
\end{figure}

\begin{figure}
    \centering
   \includegraphics[width=\linewidth]{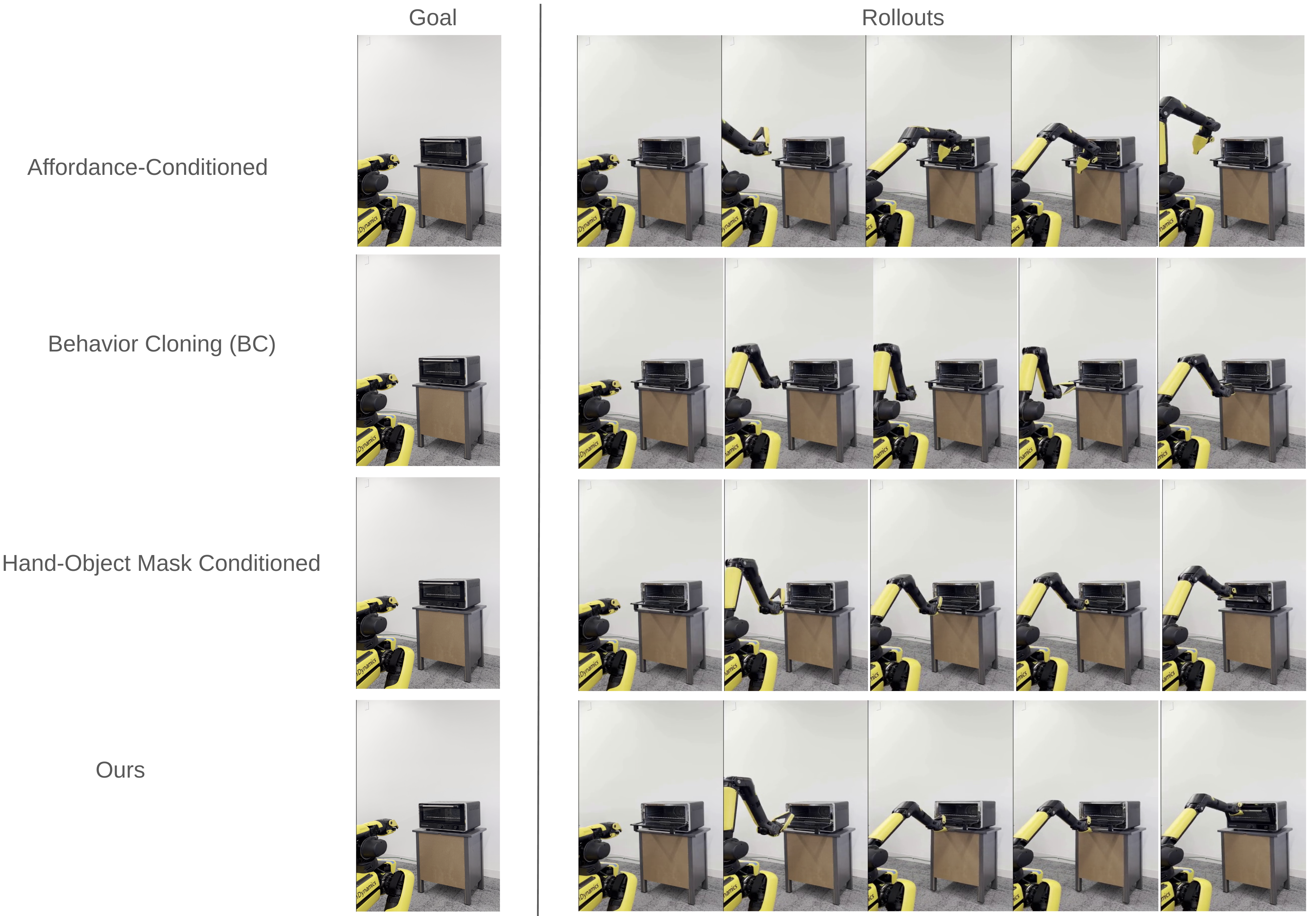}
    \caption{Compositional Generalization (CG). We show rollouts from baselines for the same goal. The views are from a third person camera.}
    \label{fig:enter-label}
\end{figure}
\begin{figure}
    \centering
   \includegraphics[width=\linewidth]{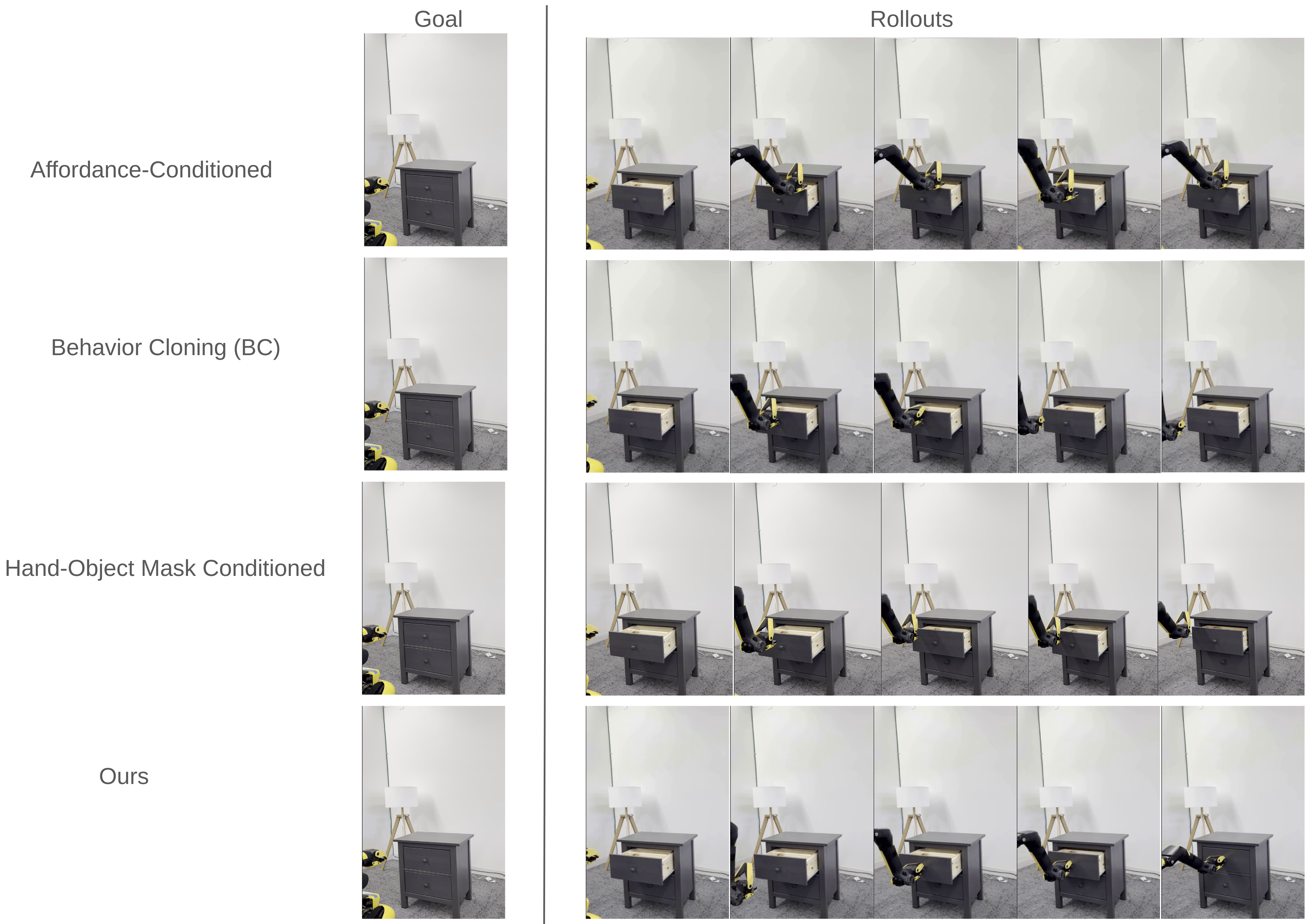}
    \caption{Standard Generalization (G). We show rollouts from baselines for the same goal. The views are from a third person camera.}
    \label{fig:enter-label}
\end{figure}
\begin{figure}
    \centering
   \includegraphics[width=\linewidth]{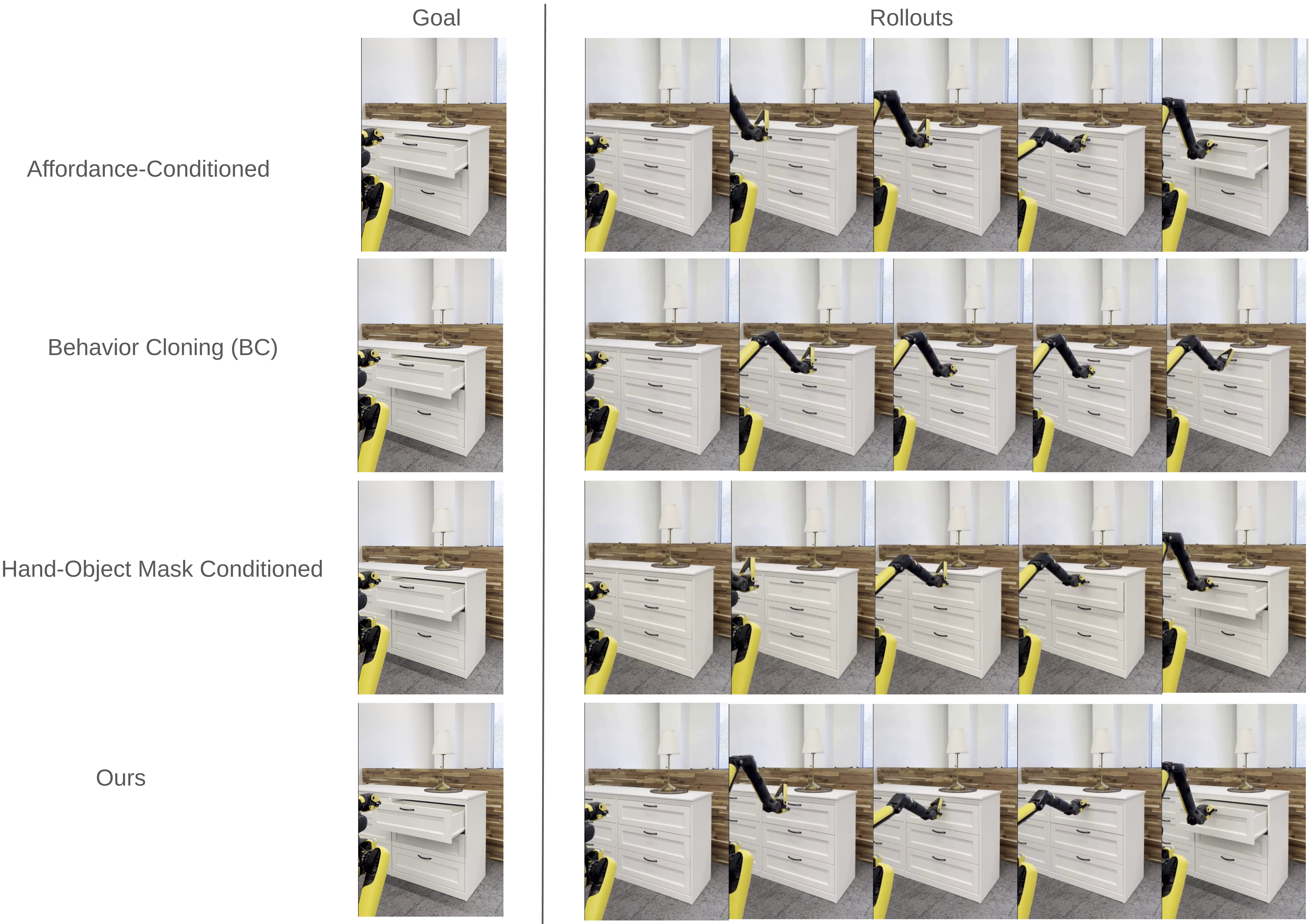}
    \caption{Mild Generalization (MG). We show rollouts from baselines for the same goal. The views are from a third person camera.}
    \label{fig:enter-label}
\end{figure}
\begin{figure}
    \centering
   \includegraphics[width=\linewidth]{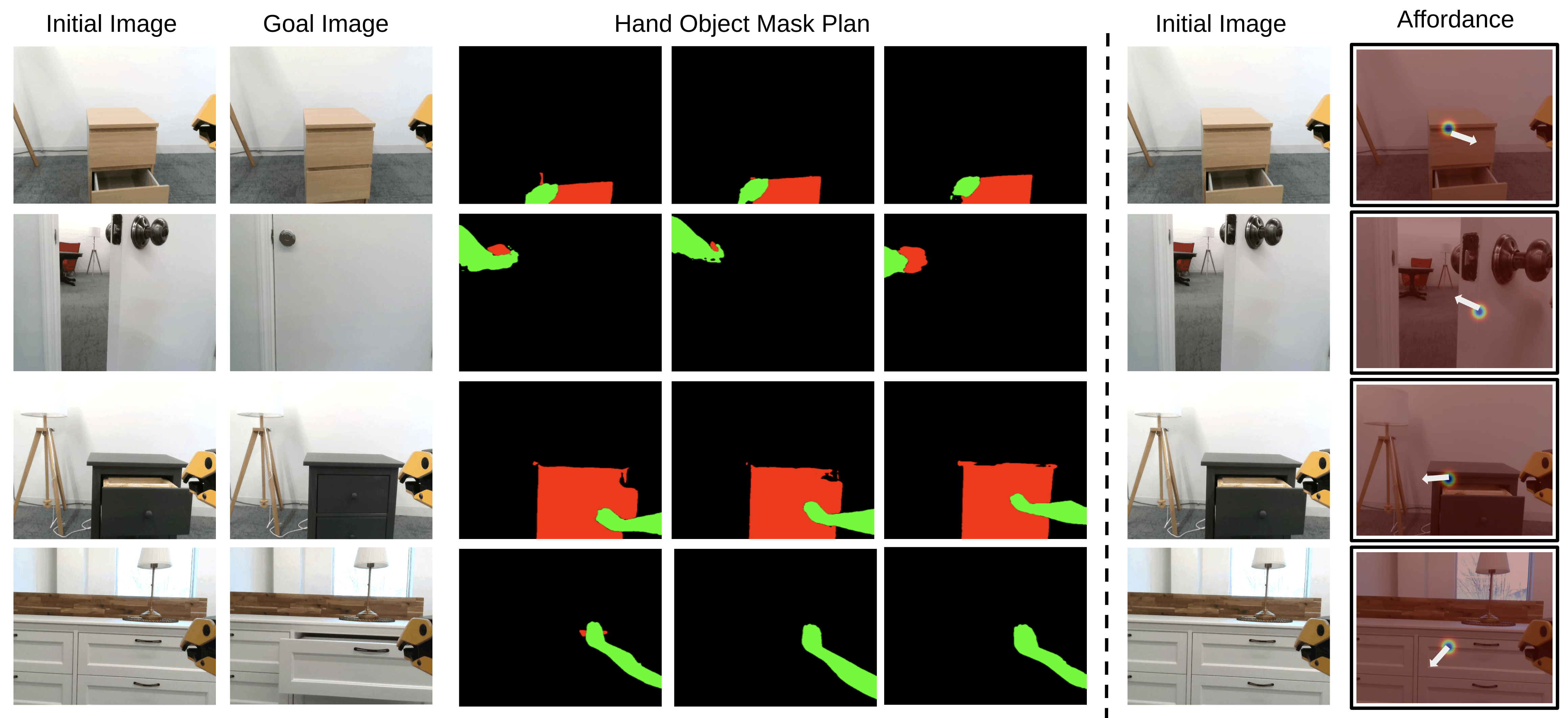}
    \caption{We show visualizations of predictions from the Hand-Object Mask Prediction and Affordance Prediction baselines, on different initial and goal images in the robot's environment.}
    \label{fig:enter-label}
\end{figure}
\end{document}